%% file: span_copying.tex
\def\year{2021}\relax
\documentclass[letterpaper]{article} 
\usepackage{aaai21}  
\usepackage{times}  
\usepackage{helvet} 
\usepackage{courier}  
\usepackage[hyphens]{url}  
\usepackage{graphicx} 
\usepackage{natbib}  
\usepackage{caption} 
\usepackage[switch]{lineno}
\usepackage[paperVersion]{marcpaper}

\newcommand*\patchAmsMathEnvironmentForLineno[1]{%
  \expandafter\let\csname old#1\expandafter\endcsname\csname #1\endcsname
  \expandafter\let\csname oldend#1\expandafter\endcsname\csname end#1\endcsname
  \renewenvironment{#1}%
     {\linenomath\csname old#1\endcsname}%
     {\csname oldend#1\endcsname\endlinenomath}}%
\newcommand*\patchBothAmsMathEnvironmentsForLineno[1]{%
  \patchAmsMathEnvironmentForLineno{#1}%
  \patchAmsMathEnvironmentForLineno{#1*}}%
\AtBeginDocument{%
\patchBothAmsMathEnvironmentsForLineno{equation}%
\patchBothAmsMathEnvironmentsForLineno{align}%
\patchBothAmsMathEnvironmentsForLineno{flalign}%
\patchBothAmsMathEnvironmentsForLineno{alignat}%
\patchBothAmsMathEnvironmentsForLineno{gather}%
\patchBothAmsMathEnvironmentsForLineno{multline}%
}

\newcommand{\genTok}[1]{\ensuremath{\mathsf{Gen}(#1)}\xspace}
\newcommand{\copyTok}[1]{\ensuremath{\mathsf{Copy}(#1)}\xspace}
\newcommand{\copySpan}[2]{\ensuremath{\mathsf{Copy}(#1:#2)}\xspace}
\newcommand{\evalAct}[1]{\ensuremath{\llbracket #1 \rrbracket}\xspace}
\newcommand{\actLen}[1]{\ensuremath{\lvert #1 \rvert}}
\newcommand{\inTok}[0]{\ensuremath{\mathit{in}}}
\newcommand{\inRepr}[0]{\ensuremath{\vect{r}}}
\newcommand{\outTok}[0]{\ensuremath{o}}
\newcommand{\outTokSlice}[1]{\ensuremath{\vect{\outTok}_{[#1]}}}

\newcommand{\seqSimpleCopy}{\textsc{Seq2Seq+CopyTok}\xspace}
\newcommand{\seqCopySpan}{\textsc{Seq2Seq+CopySpan}\xspace}

\title{Copy That! Editing Sequences by Copying Spans}
\author{
    Sheena Panthaplackel,\textsuperscript{\rm 1}\thanks{Work done during an internship in Microsoft Research, Cambridge, UK.}
    Miltiadis Allamanis,\textsuperscript{\rm 2}
    Marc Brockschmidt\textsuperscript{\rm 2}
    \\
}
\affiliations{

    \textsuperscript{\rm 1}
    University of Texas at Austin, Texas, USA
    spantha@cs.utexas.edu\\
    
    \textsuperscript{\rm 2}
    Microsoft Research, Cambridge, UK
    \{miallama, mabrocks\}@microsoft.com\\

}

\begin{document}
\maketitle

\begin{abstract}
   Neural sequence-to-sequence models are finding increasing use in editing
   of documents, for example in correcting a text document or repairing source
   code.
   In this paper, we argue that common seq2seq models (with a facility to copy
   single tokens) are \emph{not} a natural fit for such tasks, as they have to
   explicitly copy each unchanged token.
   We present an extension of seq2seq models capable of copying entire spans of
   the input to the output in one step, greatly reducing the number of decisions
   required during inference.
   This extension means that there are now many ways of generating the same
   output, which we handle by deriving a new objective for training and a variation
   of beam search for inference that explicitly handles this problem.

   In our experiments on a range of editing tasks of natural language and source
   code, we show that our new model consistently outperforms simpler baselines.
\end{abstract}

\newcommand{\bfpsmall}{BFP$_{\text{small}}$\xspace}
\newcommand{\bfpmedium}{BFP$_{\text{medium}}$\xspace}

\section{Introduction}
\label{sect:introduction}
\input{text/introduction}

\section{Model}
\label{sect:model}
\input{text/model}

\section{Related Work}
\label{sect:relwork}
\input{text/related_work}

\section{Experimental Evaluation}
\label{sect:experiments}
\input{text/experiments}

\section{Conclusion}
\label{sect:conclusion}
\input{text/conclusion}

\newpage
\bibliography{bibliography}

\newpage
\appendix
\label{sect:appendix}
\input{text/appendix}

\end{document}

%% file: text/introduction.tex
Intelligent systems that \emph{assist} users in achieving their goals have
become a focus of recent research.
One class of such systems are intelligent editors that identify and correct
errors in documents while they are written.
Such systems are usually built on the seq2seq~\citep{sutskever2014sequence}
framework, in which an input sequence (the current state of the document)
is first encoded into a vector representation and a decoder then constructs
a new sequence from this information.
Many applications of the seq2seq framework require the decoder to copy some
words in the input.
An example is machine translation, in which most words are generated,
but rare elements such as names are copied from the input.
This can be implemented in an elegant manner by equipping the decoder with
a facility that can ``point'' to words from the input, which are then copied
into the output~\citep{vinyals2015pointer,grave2016improving,gulcehre2016pointing,merity2016pointer}.

\emph{Editing} sequences poses a different problem from other seq2seq
tasks, as in many cases, \emph{most} of the input remains unchanged and needs
to be reproduced.
When using existing decoders, this requires painstaking word-by-word copying of
the input.
In this paper, we propose to extend a decoder with a facility to copy entire
spans of the input to the output in a single step, thus greatly reducing the
number of decoder steps required to generate an output.
This is illustrated in \rF{fig:wikiedits sample}, where we show how our
model inserts two new words into a sentence by copying two spans of (more than)
twenty tokens each.

However, this decoder extension exacerbates a well-known problem in training decoders
with a copying facility: a target sequence can be generated in many
different ways when an output token can be generated by different means.
In our setting, a sequence of tokens can be copied token-by-token, in pairs of
tokens, in triplets, \etc or in just a single step.
In practice, we are interested in copying the longest spans possible, as copying
longer spans both speeds up decoding at inference time and reduces the potential
for making mistakes.
To this end, we derive a training objective that marginalises over
all different generation sequences yielding the correct output, which implicitly
encourages copying longer spans.
At inference time, we solve this problem by a variation of beam search that
``merges'' rays in the beam that generate the same output by different means.

\begin{figure*}
    \input{figures/editexample.tex}
    \caption{Sample edit generated by our span-copying model on the WikiAtomicEdits dataset
            on the edit representation task of \citet{yin2019learning}.
            $\vartriangleright$ and $\vartriangleleft$ represent the BPE
            start/end tokens. The model first copies a long initial
            span of text \copySpan{1}{28}. The next two actions generate the
            tokens ``and'' and ``translator'', as in a standard sequence generation models.
            Then, the model again copies a long span of text and finally
            generates the end-of-sentence token (not shown).}\label{fig:wikiedits sample}
\end{figure*}

In summary, this paper
 (i) introduces a sequence decoder able to copy entire spans (\rSC{sect:model});
 (ii) derives a training objective that encourages our decoder to copy \emph{long}
  spans when possible;
 (iii) discusses a variation of beam search method matching our new training objective;
 and
 (iv) presents extensive experiments showing that the span-copying decoder improves
  on editing tasks on natural language and program source code (\rSC{sect:experiments}).

%% file: figures/editexample.tex
\newcommand{\ann}[1]{\tikz[overlay,remember picture,baseline=(#1.base)] \node (#1) {\color{white}};}
\newcommand{\bpestart}{{\tiny$^\vartriangleright$}}
\newcommand{\bpeend}{{\tiny$^\vartriangleleft$}~}
\begin{minipage}{0.43\textwidth}
\textbf{Input}

\myhighlight[blue!8,draw=blue]{charles \bpestart dea lt ry \bpeend \bpestart loc ock \bpeend ( september 27 , 1862}\ann{span1} \myhighlight[blue!8,draw=blue]{\;--- may 13 , 1946 ) , was a british  literary scholar }
\myhighlight[yellow!8,draw=yellow]{, who wrote on a wide array of subjects , including}\ann{span3} \myhighlight[yellow!8,draw=yellow]{chess , billiards and \bpestart cro que t \bpeend .}
\end{minipage}
\hfill
\begin{minipage}{0.43\textwidth}
\textbf{Output}

\ann{span2}\myhighlight[blue!8,draw=blue]{charles \bpestart dea lt ry \bpeend \bpestart loc ock \bpeend ( september 27 , 1862} \myhighlight[blue!8,draw=blue]{\;--- may 13 , 1946 ) , was a british literary scholar }
 and translator \myhighlight[yellow!8,draw=yellow]{, who wrote on a wide array of subjects ,} \ann{span4}\myhighlight[yellow!8,draw=yellow]{including chess , billiards and \bpestart cro que t \bpeend .}
\end{minipage}

\begin{tikzpicture}[overlay,remember picture]
    \draw[<-, blue] ($(span1) + (0,0.2)$) -- ($(span2) + (0,-0.2)$)  node [pos=0.5,above] {\scriptsize $a_1$: \copySpan{1}{28}};
    \draw[<-, black!80!yellow] (span3) -- (span4) node [pos=0.5,above] {\scriptsize $a_4$: \copySpan{28}{48}};
\end{tikzpicture}


%% file: text/model.tex
The core of our new decoder is a span-copying mechanism that can be viewed as a
generalisation of pointer networks used for copying single tokens~\citep{vinyals2015pointer,grave2016improving,gulcehre2016pointing,merity2016pointer}.
Concretely, modern sequence decoders treat copying from the input sequence as an
alternative to generating a token from the decoder vocabulary, \ie at each
step, the decoder can either generate a token $t$ from its vocabulary or it
can copy the $i$-th token of the input.
We view these as potential \emph{actions} the decoder can perform and denote
them by \genTok{t} and \copyTok{i}.

Formally, given an input sequence $\vect{\inTok} = \inTok_0 \ldots \inTok_{n-1}$, the probability of
a target sequence $\vect{\outTok} = \outTok_0 \ldots \outTok_{m-1}$ is commonly factorised into sequentially
generating all tokens of the output.
\begin{align}
    p(\outTokSlice{:m} \mid \vect{\inTok}) =
        \prod_{0 \leq j < m} p(\outTok_j \mid \vect{\inTok}, \outTokSlice{:j} ) \label{eq:seq-prob1}
\end{align}
Here, $\outTokSlice{:j}$ denotes the output tokens $\outTok_0 \ldots \outTok_{j-1}$,
following Python's slicing notation.
$p(\outTok_j \mid \vect{\inTok}, \outTokSlice{:j})$ is the probability of generating
the token $\outTok_j$, which is simply the probability of the $\genTok{t}$ action
in the absence of a copying mechanism.\footnote{Note that all occurrences of $p$ (and $q$ below)
 are implicitly (also) conditioned on the input sequence $\vect{\inTok}$, and so we drop
 this in the following to improve readability.}
When we can additionally copy tokens from the input, this probability is the sum
of probabilities of all correct actions.
To formalise this, we denote evaluation of an action $a$ into a concrete token as
\evalAct{a}, where $\evalAct{\genTok{t}} = t$ and $\evalAct{\copyTok{i}} = \inTok_i$.
Using $q(a \mid \vect{\outTok})$ to denote the probability of emitting an action $a$
after generating the partial output $\vect{\outTok}$, we complete
Eq. \rEq{eq:seq-prob1} by defining
\begin{align*}
    p(\outTok_j \mid \outTokSlice{:j}) = \sum_{a, \evalAct{a} = \outTok_j} q(a \mid \outTokSlice{:j}),
\end{align*}
\ie the sum of the probabilities of all correct actions.

\paragraph{Modelling Span Copying}
In this work, we are interested in copying whole subsequences of the input,
introducing a sequence copying action \copySpan{i}{j} with
 $\evalAct{\copySpan{i}{j}} = \inTok_i \ldots \inTok_{j-1}$.
This creates a problem because the number of actions required to generate
an output token sequence is not equal to the length of the output sequence
anymore; indeed, there may be many action sequences of different length that
can produce the correct output.

\tikzset{outSeq/.style={action/.style={text=black, anchor=north west},
                        seqProb/.style={rectangle, color=black, fill=gray!25, draw}
                        }
        }
\def\ActionDist{.25}
\def\ProbDist{2}
\def\ProbInnerSep{0.1}
\tikzset{>=stealth'}
\pgfdeclarelayer{background}
\pgfsetlayers{background,main}
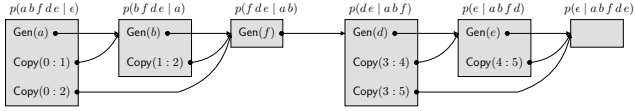
\begin{figure}
  \resizebox{\columnwidth}{!}{%
    \begin{tikzpicture}[outSeq]
        \node (11-genA) [action] at (0, 0)
            {\genTok{a}};
        \node (12-copy01) [action] at ($(11-genA.south west) + (0, -\ActionDist)$)
            {\copySpan{0}{1}};
        \node (13-copy02) [action] at ($(12-copy01.south west) + (0, -\ActionDist)$)
            {\copySpan{0}{2}};
        \begin{pgfonlayer}{background}
            \path[seqProb]
                ($(11-genA.north west) + (-\ProbInnerSep, \ProbInnerSep)$)
                rectangle
                ($(13-copy02.south east) + (\ProbInnerSep, -\ProbInnerSep)$);
        \end{pgfonlayer}
        \node (probABFDE-label) [anchor=south west] at ($(11-genA.north west) + (-\ProbInnerSep, 1.2*\ProbInnerSep)$)
            {$p(a \, b \, f \, d \, e \mid \epsilon)$};

        \node (21-genB) [action] at ($(11-genA.north east) + (\ProbDist, 0)$)
            {\genTok{b}};
        \node (22-copy12) [action] at ($(21-genB.south west) + (0, -\ActionDist)$)
            {\copySpan{1}{2}};
        \begin{pgfonlayer}{background}
            \path[seqProb]
                ($(21-genB.north west) + (-\ProbInnerSep, \ProbInnerSep)$)
                rectangle
                ($(22-copy12.south east) + (\ProbInnerSep, -\ProbInnerSep)$);
        \end{pgfonlayer}
        \node (probBFDE-label) [anchor=south west] at ($(21-genB.north west) + (-\ProbInnerSep, 1.2*\ProbInnerSep)$)
            {$p(b \, f \, d \, e \mid a)$};

        \node (31-genF) [action] at ($(21-genB.north east) + (\ProbDist, 0)$)
            {\genTok{f}};
        \begin{pgfonlayer}{background}
            \path[seqProb]
                ($(31-genF.north west) + (-\ProbInnerSep, \ProbInnerSep)$)
                rectangle
                ($(31-genF.south east) + (\ProbInnerSep, -\ProbInnerSep)$);
        \end{pgfonlayer}
        \node (probFDE-label) [anchor=south west] at ($(31-genF.north west) + (-\ProbInnerSep, 1.2*\ProbInnerSep)$)
            {$p(f \, d \, e \mid a \, b)$};

        \node (41-genD) [action] at ($(31-genF.north east) + (\ProbDist, 0)$)
            {\genTok{d}};
        \node (42-copy34) [action] at ($(41-genD.south west) + (0, -\ActionDist)$)
            {\copySpan{3}{4}};
        \node (43-copy35) [action] at ($(42-copy34.south west) + (0, -\ActionDist)$)
            {\copySpan{3}{5}};
        \begin{pgfonlayer}{background}
            \path[seqProb]
                ($(41-genD.north west) + (-\ProbInnerSep, \ProbInnerSep)$)
                rectangle
                ($(43-copy35.south east) + (\ProbInnerSep, -\ProbInnerSep)$);
        \end{pgfonlayer}
        \node (probDE-label) [anchor=south west] at ($(41-genD.north west) + (-\ProbInnerSep, 1.2*\ProbInnerSep)$)
            {$p(d \, e \mid a \, b \, f)$};

        \node (51-genE) [action] at ($(41-genD.north east) + (\ProbDist, 0)$)
            {\genTok{e}};
        \node (52-copy45) [action] at ($(51-genE.south west) + (0, -\ActionDist)$)
            {\copySpan{4}{5}};
        \begin{pgfonlayer}{background}
            \path[seqProb]
                ($(51-genE.north west) + (-\ProbInnerSep, \ProbInnerSep)$)
                rectangle
                ($(52-copy45.south east) + (\ProbInnerSep, -\ProbInnerSep)$);
        \end{pgfonlayer}
        \node (probE-label) [anchor=south west] at ($(51-genE.north west) + (-\ProbInnerSep, 1.2*\ProbInnerSep)$)
            {$p(e \mid a \, b \, f \, d)$};

        \node (61-eps) [action] at ($(51-genE.north east) + (\ProbDist, 0)$)
            {\phantom{\genTok{\epsilon}}};

        \begin{pgfonlayer}{background}
            \path[seqProb]
                ($(61-eps.north west) + (-\ProbInnerSep, \ProbInnerSep)$)
                rectangle
                ($(61-eps.south east) + (\ProbInnerSep, -\ProbInnerSep)$);
        \end{pgfonlayer}
        \node (probEps-label) [anchor=south west] at ($(61-eps.north west) + (-4*\ProbInnerSep, 1.2*\ProbInnerSep)$)
            {$p(\epsilon \mid a \, b \, f \, d \, e)$};

        \path[{Circle[fill=black]}->]
            (11-genA.east) edge ($(21-genB.west) + (-\ProbInnerSep, 0)$)
            (12-copy01.east) edge[bend right] ($(21-genB.west) + (-\ProbInnerSep, 0)$)
            (13-copy02.east) edge[{Circle[fill=black]}-] ($(13-copy02.east) + (1.5*\ProbDist, 0)$)
            ($(13-copy02.east) + (1.5*\ProbDist, 0)$) edge[bend right,->] ($(31-genF.west) + (-\ProbInnerSep, 0)$)

            (21-genB.east) edge ($(31-genF.west) + (-\ProbInnerSep, 0)$)
            (22-copy12.east) edge[bend right] ($(31-genF.west) + (-\ProbInnerSep, 0)$)

            (31-genF.east) edge ($(41-genD.west) + (-\ProbInnerSep, 0)$)

            (41-genD.east) edge ($(51-genE.west) + (-\ProbInnerSep, 0)$)
            (42-copy34.east) edge[bend right] ($(51-genE.west) + (-\ProbInnerSep, 0)$)
            (43-copy35.east) edge[{Circle[fill=black]}-] ($(43-copy35.east) + (1.5*\ProbDist, 0)$)
            ($(43-copy35.east) + (1.5*\ProbDist, 0)$) edge[bend right,->] ($(61-eps.west) + (-\ProbInnerSep, 0)$)

            (51-genE.east) edge ($(61-eps.west) + (-\ProbInnerSep, 0)$)
            (52-copy45.east) edge[bend right] ($(61-eps.west) + (-\ProbInnerSep, 0)$)
            ;
    \end{tikzpicture}%
  }
  \caption{\label{fig:copyEx}Illustration of different ways of generating the
   sequence
    $a \, b \, f \, d \, e$
   given an input of
    $a \, b \, c \, d \, e$.
   Each box lists all correct actions at a given point in the generation process,
   and the edges after an action indicate which suffix token sequence still needs
   to be generated after it.
   We use $\epsilon$ to denote the empty sequence, either as prefix or suffix.}
\end{figure}
As an example, consider \rF{fig:copyEx}, which illustrates all action sequences
generating the output $a \, b \, f \, d \, e$ given the input $a \, b \, c \, d \, e$.
For example, we can initially generate the token $a$, or copy it from the input,
or copy the first two tokens.
If we choose one of the first two actions, we then have the choice of either generating
the token $b$ or copying it from the input and then have to generate the token
$f$.
Alternatively, if we initially choose to copy the first two tokens, we have to
generate the token $f$ next.
We can compute the probability of generating the target sequence by traversing
the diagram from the right to the left.
$p(\epsilon \mid a \, b \, f \, d \, e)$ is simply the probability of emitting
a stop token and requires no recursion, where $\epsilon$ denotes the empty
sequence.
$p(e \mid a \, b \, f \, d)$ is the sum of the probabilities
 $q(\genTok{e} \mid a \, b \, f \, d) \cdot p(\epsilon \mid a \, b \, f \, d \, e)$
and
 $q(\copySpan{4}{5} \mid a \, b \, f \, d) \cdot p(\epsilon \mid a \, b \, f \, d \, e)$,
which re-use the term we already computed.
Following this strategy, we can compute the probability of generating the
output token sequence by computing probabilities of increasing longer suffixes
of it (essentially traversing the diagram in \rF{fig:copyEx} from right to left).

Formally, we reformulate Eq. \rEq{eq:seq-prob1} into a recursive definition that
marginalises over all different sequences of actions generating the correct
output sequence, following the strategy illustrated in \rF{fig:copyEx}.
For this we define \actLen{a}, the length of the output of an action, \ie,
$\actLen{\genTok{t}} = 1$ and $\actLen{\copySpan{i}{j}} = j - i$.
Note that w.l.o.g., we assume that actions \copySpan{i}{j} with $j \leq i$
do \emph{not} exist, \ie copies of zero-length spans are explicitly ignored.
\begin{align}
    p(\outTokSlice{k:} \mid \outTokSlice{:k}) =
    \hspace{-2ex}\sum_{\substack{a, \exists \ell . \actLen{a} = \ell\\
                    \evalAct{a} = \outTokSlice{k:k+\ell}}}\hspace{-2ex}
        q(a \mid \outTokSlice{:k}) \cdot p(\outTokSlice{k+\ell:} \mid \outTokSlice{:k+\ell})
        \label{eq:seq-prob2}
\end{align}
Here, the probability of generating the correct suffix
is only conditioned on the sequence generated so far and \emph{not} on the
concrete actions that yielded it.
In practice, we implement this by conditioning our model of $q$ at
timestep $k$ on a representation $\vect{h}_k$ computed from the
partial output sequence $\vect{\outTok}[:k]$.
In RNNs, this is modelled by feeding the sequence of emitted tokens into the
decoder, no matter how the decoder determined to emit these, and thus, one
\copySpan{i}{j} action may cause the decoder RNN to take multiple timesteps
to process the copied token sequence.
In causal self-attentional settings, this is simply the default behaviour.
Finally, note that for numerical stability purposes our implementation
works on the log-probability space as it is common for such methods,
implementing the summation of probabilities with the standard log-sum-exp trick.

\paragraph{Modelling Action Choices}
It remains to explain how we model the per-step action distribution
$q(a \mid \vect{\outTok})$.
We assume that we have per-token encoder representations $\inRepr_0 \ldots
\inRepr_{n-1}$ of all input tokens and a decoder state $\vect{h}_k$ obtained after
emitting the prefix \outTokSlice{:k}.
This can be the state of an RNN cell after processing the sequence \outTokSlice{:k}
(potentially with attention over the input) or the representation of a
self-attentional model processing that sequence.

As in standard sequence decoders, we use an output embedding projection
applied to $\vect{h}_k$ to obtain scores $s_{k,v}$ for all tokens in the
decoder vocabulary.
To compute a score for a \copySpan{i}{j} action, we use a linear layer
$\vect{W}$ to project the concatenation $\inRepr_i \Vert \inRepr_{j-1}$ of the
(contextualised) embeddings of the respective input tokens to the same
dimension as $\vect{h}_k$ and then compute their inner product:
\begin{align*}
    s_{k, [i:j]} = (\vect{W} \cdot (\inRepr_i \Vert \inRepr_{j-1})) \cdot \vect{h}_k^\top
\end{align*}
We then concatenate all $s_{k, v}$ and $s_{k, [i:j]}$ and apply a softmax
to obtain our action distribution $q(a \mid \vect{\outTok})$. Note that
for efficient computation in GPUs, we compute the $s_{k, [i:j]}$
for all $i$ and $j$ and mask all entries where $j \leq i$.

\paragraph{Training Objective}
We train in the standard teacher-forcing supervised sequence decoding setting, feeding to the
decoder the correct output sequence independent of its decisions.
We train by maximising $p(\vect{\outTok} \mid \epsilon)$ unrolled
according to Eq. \rEq{eq:seq-prob2}.
One special case to note is that we make a minor but important modification to handle
generation of out-of-vocabulary words: \emph{iff} the correct token can
be copied from the input, \genTok{\mathsf{UNK}} is considered to be an
incorrect action; otherwise only \genTok{\mathsf{UNK}} is correct.
This is necessary to avoid pathological cases in which there is no action
sequence to generate the target sequence correctly.

We found that using the marginalisation in Eq. \rEq{eq:seq-prob2}
during training is crucial for good results.
Our experiments (cf. \rSC{sect:correction_experiments}) include an ablation in
which we generate a per-token
loss based only on the correct actions at each output token, without taking the
remainder of the sequence into account (\ie, at each point in time, we used a
``multi-hot'' objective in which the loss encourages picking any one of the
correct actions).
Training using this objective yielded a decoder which would most often only copy
sequences of length one.
In contrast, our marginalised objective penalises long sequences of actions and
hence pushes the model towards copying longer spans when possible.
The reason for this is that constructing an output sequence from longer spans
implies that the required action sequence is shorter.
As each action decision ($q$ in Eq.~\rEq{eq:seq-prob2}) 
``costs'' some probability mass, as in practice $q$ will assign some probability
to incorrect choices.
The marginalization in our objective ensures that the model is rewarded for
preferring to copy longer (correct) spans,
which imply shorter action sequences, and hence fewer places at which probability mass is ``spent''.
At the same time, we do not force the model to copy the longest possible sequence but
instead allow the optimization process to find the best trade-off.

\paragraph{Beam Decoding}
Our approach to efficiently evaluate Eq. \rEq{eq:seq-prob2} at training time
relies on knowledge of the ground truth sequence and so we need to employ
another approach at inference time.
We use a variation of standard beam search which handles the fact that action
sequences of the same length can lead to sequences of different lengths.
For this, we consider a forward version of Eq. \rEq{eq:seq-prob2}
in which we assume to have a set of action sequences $\mathcal{A}$ and
compute a lower bound on the true probability of a sequence $\vect{\outTok}$
by considering all action sequences in $\mathcal{A}$ that evaluate
to $\outTok_0 \ldots \outTok_{k-1}$:
\begin{align}
    p(\outTokSlice{:k}) &
    \geq \hspace{-3ex}\sum_{\substack{[a_1 \ldots a_n] \in \mathcal{A}\\
                      \evalAct{a_1} \Vert \ldots \Vert \evalAct{a_n} = \outTokSlice{:k}}}\hspace{-1ex}
        \prod_{1 \leq i \leq n} q(a_i \mid \evalAct{a_1} \Vert \ldots \Vert \evalAct{a_{i-1}})
        \label{eq:seq-prob3}
\end{align}
If $\mathcal{A}$ contains the set of all action sequences generating the output
sequence $\outTok_0 \ldots \outTok_{k-1}$, Eq. \rEq{eq:seq-prob3} is an equality.
At inference time, we under-approximate $\mathcal{A}$ by generating likely
action sequences using beam search.
However, we have to explicitly implement the summation of the probabilities
of action sequences yielding the same output sequence.
This could be achieved by a final post-processing step (as in Eq. \rEq{eq:seq-prob3}),
but we found that it is more effective to ``merge'' rays generating the same
sequence during the search.
In the example shown in \rF{fig:copyEx}, this means to sum up the probabilities
of (for example) the action sequences $\genTok{a} \genTok{b}$ and $\copySpan{0}{2}$,
as they generate the same output.
To group action sequences of different lengths, our search
procedure is explicitly considering the length of the generated token sequence
and ``pauses'' the expansion of action sequences that have generated longer
outputs.
The pseudocode for this procedure is shown in \rA{alg:beam-search}, where 
\lstinline{group_by_toks} merges different rays generating the same output.

\begin{algorithm}[t]
    \begin{lstlisting}
def beam_search(beam_size)
  beam = [ {toks: [START_OF_SEQ], prob: 1} ]
  out_length = 1
  while unfinished_rays(beam):
    new_rays = []
    for ray in beam:
      if len(ray.toks) > out_length
        or ray.toks[-1] == END_OF_SEQ:
        new_rays.append(ray)
      else:
        for (act, act_prob) in (*$q(\cdot \mid$*) ray.toks(*$)$*):
          new_rays.append(
            {toks: ray.toks(*$\ \Vert \ \llbracket$*)act(*$\rrbracket$*),
             prob: ray.prob*act_prob})
    beam = top_k(group_by_toks(new_rays),
                     k=beam_size)
    out_length += 1
  return beam
    \end{lstlisting}
    \caption{\label{alg:beam-search}Python-like pseudocode of beam search for span-copying decoders.}
\end{algorithm}

\paragraph{Complexity}
Our objective in Eq.~\rEq{eq:seq-prob2} can be computed using the described dynamic program with complexity in $O(N^2)$,
where $N$ is the sequence length. 
On a GPU, this can be efficiently parallelised, such that for all reasonable sequence lengths, only a linear number of (highly parallel) operations is required.
In practice, a slowdown is only observed during training, but not during beam decoding.
For example, during training for the task \bfpsmall (see \rSC{sect:experiments}),
computing the per-step log-probabilities takes about 80ms per minibatch, whereas marginalisation takes about 52ms per minibatch.
This constitutes 65\% extra time required for the marginalisation.

%% file: text/related_work.tex
Copying mechanisms are common in neural natural language processing.
Starting from pointer networks~\citep{vinyals2015pointer}, such mechanisms
have been used across a variety of domains~\citep{allamanis2016convolutional,gu2016incorporating,see2017get}
as a way to copy elements from the input to the output, usually as a way to
alleviate issues around rare, out-of-vocabulary tokens such as names.
Marginalising over a single token-copying action and a generation action
has been previously considered~\citep{allamanis2016convolutional,ling2016latent}
but these works do not consider spans longer than one ``unit''.

Most similar to our work, \citet{zhou2018sequential} propose a method to copy
spans (for text summarization tasks) by predicting the start and end of a
span to copy.
However, they do not handle the issue of different generation strategies
for the same output sequence explicitly and do not present an equivalent
to our training objective and modified beam search.
Dependent on the choice of ``copied spans'' used to train the model, it either
corresponds to the case of training our method without any marginalisation,
or one in which only one choice (such as copying the longest matching span)
is considered.
In our experiments in \rSC{sect:experiments}, we show that both variants
perform substantially less well than our marginalised objective.

Our method is somewhat related to the work of \citet{van2017multiscale,grave2019training}, who consider
``multiscale'' generation of sequences using a vocabulary of potentially overlapping
word fragments.
Doing this also requires to marginalise out different decoder actions that yield
the same output: in their case, generating a sequence from different combinations of
word fragments, in contrast to our problem of generating a sequence token-by-token
or copying a span.
Hence, their training objective is similar to our objective in Eq. \rEq{eq:seq-prob2}.
A more important difference is that they use a standard autoregressive decoder
in which the emitted word fragments are fed back as inputs.
This creates the problem of having different decoder states for the same
output sequence (dependent on its decomposition into word fragments), which
\citet{van2017multiscale} resolve by averaging the states of the decoder (an
RNN using LSTM cells).
Instead, we are following the idea of the graph generation strategy of
\citet{liu2018constrained}, where a graph decoder is only conditioned on the
partial graph that is being extended, and not the sequence of actions that
generated the graph.

Recently, a number of approaches to sequence generation avoiding the left-to-right
paradigm have been proposed~\citep{welleck2019non,stern2019insertion,gu2019levenshtein,lee2018deterministic},
usually by considering the sequence generation problem as an iterative refinement
procedure that changes or extends a full sequence in each iteration step.
Editing tasks could be handled by such models by learning to refine the input
sequence with the goal of generating the output sequence.
However, besides early experiments by \citet{gu2019levenshtein}, we are not aware
of any work that is trying to do this.
Note however that our proposed span-copying mechanism is also naturally applicable
in settings that require duplication of parts of the input, \eg when phrases
or subexpressions need to be appear several times in the output (\cf \code{obj} in
\rF{fig:code repair example} for a simple example).
Finally, sequence-refinement models could also be extended to take advantage of
our technique without large modifications, though we believe the marginalisation
over all possible insertion actions (as in Eq.~\rEq{eq:seq-prob2}) to be intractable
in this setting.
Similarly, \citet{grangier2017quickedit} present QuickEdit, a machine translation method
that accepts a source sentence (\eg in German) a guess sentence (\eg in English) that
is annotated (by humans) with change markers. It then aims to improve upon the
guess by generating a better target sentence avoiding the marked tokens.
This is markedly different as the model accepts as input the spans that need
to be removed or retained in the guess sentence. In contrast, our model needs to
automatically infer this information.

An alternative to sequence generation models for edits is the work of \citet{gupta2017deepfix},
who propose to repair source code by first pointing to a single line in the output
and then only generate a new version of that line.
However, this requires a domain-specific segmentation of the input -- lines are
often a good choice for programs, but (multi-line) statements or expressions
are valid choices as well.
Furthermore, the approach still requires to generate a sequence that is similar
to the input line and thus could profit from our span-copying approach.

%% file: text/experiments.tex
We evaluate our RNN-based implementation on two types of tasks.
First, we evaluate the performance of our models in the setting of learning
edit representations~\citep{yin2019learning} for natural language and code
changes.
Second, we consider correction-style tasks in which a model has to identify
an error in an input sequence and then generate an output sequence that
is a corrected version of the input.
In the evaluation below, \seqSimpleCopy denotes a variant of our \seqCopySpan
model in which the decoder can only copy single tokens. For all experiments,
we use a single NVidia K80 GPU.

\subsection{Edit Representations}
We first consider the task of learning edit representations~\citep{yin2019learning}.
The goal is to use an autoencoder-like model structure to
learn useful representations of edits of natural language and source code.
The model consists of an edit encoder $f_\Delta(x_-, x_+)$ to transform the
edit between $x_-$ and $x_+$ into an edit representation.
Then, a neural editor $\alpha(x_-, f_\Delta(x_-, x_+))$ uses $x_-$ and the edit
representation to reconstruct $x_+$ as accurately as possible.
We follow the same structure and employ our \seqCopySpan decoder to model
the neural editor $\alpha$.
We perform our experiments on the datasets used by \citet{yin2019learning}.

Our editor models have a 2-layer biGRU encoder with hidden size of 64,
a single layer GRU decoder with hidden size of 64, tied embedding layers
with a hidden size of 64 and use a dropout rate of 0.2. In all cases the edit encoder
$f_\Delta$ is a 2-layer biGRU with a hidden size of 64. The GRU decoders of both models
use a Luong-style attention mechanism~\citep{luong2015effective}.

\begin{table}[t]\centering
  \caption{Evaluation of models on the edit representation tasks of \citet{yin2019learning}.} \label{tbl:edit representation eval}
  \resizebox{\columnwidth}{!}{%
  \begin{tabular}{@{}lccc@{}}
    \toprule
             & WikiAtomicEdits &   \multicolumn{1}{c}{GitHubEdits} & \CSharp~Fixers \\
    \cmidrule(lr){2-2} \cmidrule(lr){3-3} \cmidrule(lr){4-4}
             & Accuracy        &     Accuracy   & Accuracy       \\
    \cmidrule(lr){2-2} \cmidrule(lr){3-3} \cmidrule(lr){4-4}
      \citet{yin2019learning}
             &          72.9\% &         59.6\% & n/a\\
      \seqSimpleCopy
             &          67.8\% &         64.4\% & 18.8\% \\
      \seqCopySpan
             &  \textbf{78.1\%}& \textbf{67.4\%}& \textbf{24.2\%} \\
    \bottomrule
  \end{tabular}
  }
\end{table}

\paragraph{Editing Wikipedia}
First, we consider the task of learning edit representations of small edits
to Wikipedia articles~\citep{faruqui2018wikiatomicedits}.\footnote{According to
  \citet{yin2019learning}, a part of the data was corrupted and hence they used a
  smaller portion of the data.}
The dataset consists of ``atomic'' edits on Wikipedia articles without
any special filters.
\rTab{tbl:edit representation eval} suggests that the span-copying model
achieves a significantly better performance in predicting the exact edit,
even though our (nominally comparable) \seqSimpleCopy
model performs worse than the model used by \citet{yin2019learning}.
Our initial example in \rF{fig:wikiedits sample} shows one edit example,
where the model, given the input text and the edit representation vector,
is able to generate the output by copying two long spans and generating only
the inserted tokens.
Note that the WikiAtomicEdits dataset is made up of only insertions and
deletions. The edit shown in \rF{fig:wikiedits sample} is generally
representative of the other edits in the test set.

\paragraph{Editing Code}
We now focus on the code editing task of \citet{yin2019learning} on the
GitHubEdits dataset, constructed from small (less than 3 lines)
code edits scraped from \CSharp~GitHub repositories. These include
bug fixes, refactorings and other code changes.
Again, the results in \rTab{tbl:edit representation eval} suggest that our
span-based models outperform the baseline by predicting the edited code more
accurately.

\citet{yin2019learning} also use the edit representations for a one-shot
learning-style task on a ``\CSharp~Fixers'' dataset, which are small changes
constructed using automatic source code rewrites.
Each edit is annotated with the used rewrite rule so that the dataset can be
used to study how well an edit representation generalises from one sample
to another.

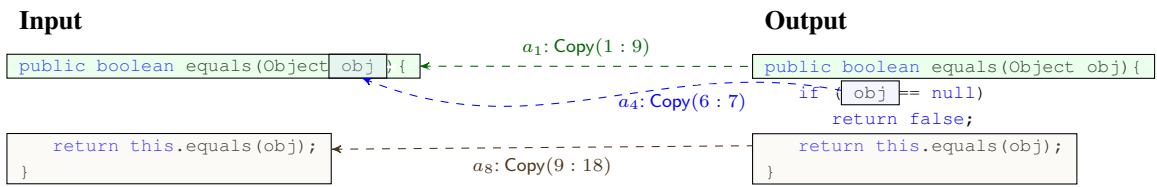
\begin{figure*}[h]
  \input{figures/repairexample}
  \caption{Generation of a test example in \bfpsmall (slightly modified for space and clarity).
      The \seqCopySpan model learns to copy long spans while generating the necessary edits.
      The non-highlighted tokens in the output are generated using \genTok{t}, whereas all
      other tokens are copied from the input.
  } \label{fig:code repair example}
\end{figure*}
As in \citet{yin2019learning}, we train the models on the larger and more general
GitHubEdits dataset.
To evaluate, we compute the edit representation $f_\Delta(x_-, x_+)$ of one sample
of a group of semantically similar edits in \CSharp~Fixers and feed it to the neural editor
with the source code of another sample, \ie, compute
$\alpha(x'_-, f_\Delta(x_-, x_+))$.
We repeat this experiment by picking the first 100 samples per fixer,
computing the edit representation of each one and applying the edit to
the other 99.
The results of this process are shown in the last column of \rTab{tbl:edit representation eval},
suggesting that our span-copying models are able to improve on the
one-shot transfer task as well.

Note that these results are not exactly comparable with
those presented in \citet{yin2019learning}, as they randomly select 10
pairs $(x_-, x_+)$, compute their edit representation and then for a given $x'_-$ compute
$\alpha(x'_-, f_\Delta(x_-, x_+))$ for each of the 10 edit representations, finally
reporting the best accuracy score among the 10 candidates.
Since this process cannot be replicated due to the random selection of samples, we instead
opted for the reproducible process described above.
\rTab{tbl:fixer eval} in appendix \ref{app:results:fixer-details} presents a breakdown of
the performance on the fixer data per fixer, showing that for some fixers our model can
substantially improve accuracy.

Overall, while our \seqSimpleCopy model is roughly on par with the numbers
reported in \citet{yin2019learning} (as expected for a re-implementation),
our new model \seqCopySpan clearly sets a new state of the art.
This improvement can be attributed to the ability of the model to copy larger
spans which allows it to better represent the relevant edits.

\subsection{Correction Tasks}
\label{sect:correction_experiments}
Correction tasks were one of the core motivations for our new decoding
strategy, as they usually require to reproduce most of the input without
changing it, whereas only few tokens are removed, updated or added.
In our experiments, we focus on corrections on source code, but also provide
an indication that our \seqCopySpan would work well for corrections
of natural language.

\paragraph{Code Repair}

Automated code repair systems~\citep{monperrus2018automatic} are commonly
composed of two components, namely a (heuristic) component that suggests
potentially fixed versions of the input, and an oracle (\eg, a harness
executing a test suite) that checks the candidates for correctness.
Recent software engineering research has started to implement the heuristic
component using seq2seq models~\citep{chen2018sequencer,tufano2019empirical,lutellier2019encore}.
The models are usually viewed as language models (conditioned on the faulty
code) or employ standard neural machine translation pipelines
mapping from ``faulty'' to ``correct'' code.
The task usually only requires minor changes to the input code and consequently
most of the input is copied into the output.
We believe that our model is a natural match for this setting.

To test this hypothesis, we use the two bug-fix pair (BFP) datasets of
\citet{tufano2019empirical}.
The \bfpsmall dataset contains pairs where each snippet has at most 50 tokens
and the \bfpmedium dataset has Java snippets containing from 50 up to 150 tokens.
In these datasets, the input is code with some form of a bug, whereas the output
is correct code.
This corpus was constructed by scraping Git commits and
filtering for those with commit messages suggesting that the edit fixes a bug.
For both the \seqSimpleCopy and \seqCopySpan models we employ a 2-layer biGRU as an
encoder and a single layer GRU decoder.
We use embeddings with 32 dimensions and GRUs with hidden units of size 128.
Note that the vocabulary size
for this task is just 400 by construction of the dataset.
We employ a Luong-style~\citep{luong2015effective} attention mechanism in the the decoders
of both models.

\begin{table*}[t]\centering
  \caption{Evaluation of models on the code repair task. Given
   an input code snippet, each model needs to predict a corrected version of that code snippet.
   ``Structural Match'' indicates that the generated output is identical to the
   target output up to renaming the identifiers (\ie, variables, functions).}\label{tbl:repairEval}
  \begin{tabular}{@{}llrrrr@{}}
    \toprule
      && Accuracy & Accuracy@20 & MRR & Structural Match \\
    \midrule
      \multicolumn{2}{l}{\underline{On \bfpsmall}}\\
      \qquad& \citet{tufano2019empirical}      &          9.2\%  &         43.5\%  &           ---  &          --- \\
      \qquad& \seqSimpleCopy                   &         14.8\%  &         42.0\%  &         0.177  &         18.2\% \\
      \qquad& \seqCopySpan                     & \textbf{17.7\%} & \textbf{45.0\%} & \textbf{0.247} & \textbf{21.2\%}\\
      \cmidrule(lr){2-6}
      \qquad& \seqCopySpan (always copy longest)&         14.2\%  &         33.7\%  &         0.174  & 14.2\%\\
      \qquad& \seqCopySpan (no marginalization) &         16.9\%  &         43.4\%  &         0.210  &   20.2\%   \\
      \multicolumn{2}{l}{\underline{On \bfpmedium}}\\
      \qquad& \citet{tufano2019empirical} &          3.2\%  &         22.2\%  &           ---  &          --- \\
      \qquad& \seqSimpleCopy              &          7.0\%  &         23.8\%  &         0.073  &          9.4\% \\
      \qquad& \seqCopySpan                &  \textbf{8.0\%} & \textbf{25.4\%} & \textbf{0.105} & \textbf{13.7\%}\\
      \cmidrule(lr){2-6}
      \qquad& \seqCopySpan (always copy longest)&     7.2\%  &         20.0\%  &         0.090  &          10.8\%\\
      \qquad& \seqCopySpan (no marginalization) &     2.5\%  &         11.1\%  &         0.035  &          3.7\%\\
    \bottomrule
  \end{tabular}
\end{table*}
\rTab{tbl:repairEval} shows the results of our models, as well as the
original results reported by \citet{tufano2019empirical}.
Overall, the \seqCopySpan model performs better on both datasets, achieving
a new state of the art.
This suggests that the span-copying mechanism is indeed beneficial in this
setting, as becomes clear in a qualitative analysis.
\rF{fig:code repair example} shows an example (slightly modified for
readability) of a code repair prediction and the span-copying actions.
In this case, the model has learned to copy all of the input code in chunks,
extending it only by inserting some new tokens in the middle.

We use this task to consider four ablations of our model, clarifying the
impact of each of the contributions of our paper.
To study the effect of marginalising over all correct choices (Eq~\rEq{eq:seq-prob2}),
we compare with two alternative solutions.
First, we train the model to always copy the longest possible span.
The results shown in \rTab{tbl:repairEval} indicate that this has a
substantial impact on results, especially for results obtained by beam search.
We believe that this is due to the fact that the model fails to capture the
entire spectrum of correct actions, as the objective penalises correct-but-not-longest
copying actions.
This leads to a lack of informed diversity, reducing the benefits of beam search.

Second, we consider an objective in which we use no marginalisation, but instead
train the model to predict any one of the correct actions at each step,
without any preference for long or short copied spans -- this corresponds to the
approach of \citet{zhou2018sequential}.
Our results show that this is competitive on shorter output sequences, but
quickly degrades for longer outputs.
We believe that this is due to the fact that the model is not encouraged to use
as few actions as possible, which consequently means that producing a correct output
can require dozens or hundreds of prediction steps.

We also evaluated our modified beam search in \rA{alg:beam-search} in comparison to
standard beam search and greedy decoding on the code repair task (see appendix
\ref{app:results:merged-beam-search} for details).
The results show (small) improvements on the quality when considering only the top
result, but substantial gains when taking more decoder results into account.

\begin{figure*}
  \begin{subfigure}[b]{0.48\textwidth}\centering
      \includegraphics[width=\textwidth]{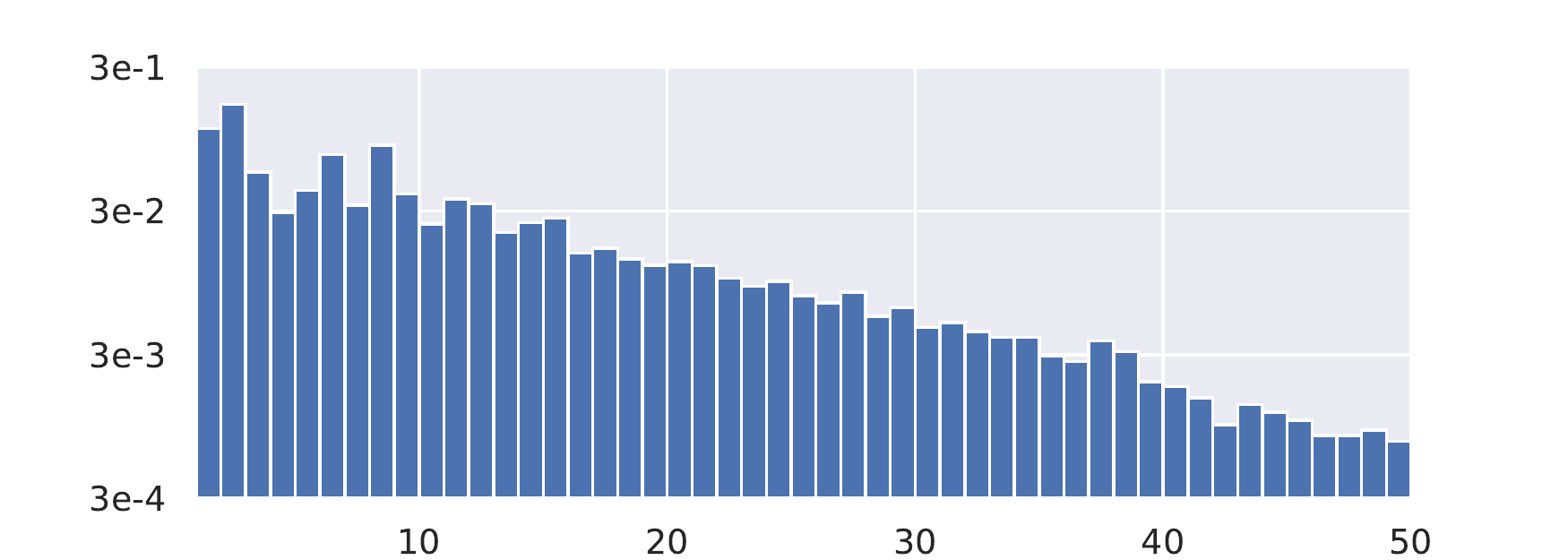}
      \caption{ \bfpsmall ($\mu=9.6$, median: 7) }
  \end{subfigure}
~
  \begin{subfigure}[b]{0.48\textwidth}\centering
      \includegraphics[width=\textwidth]{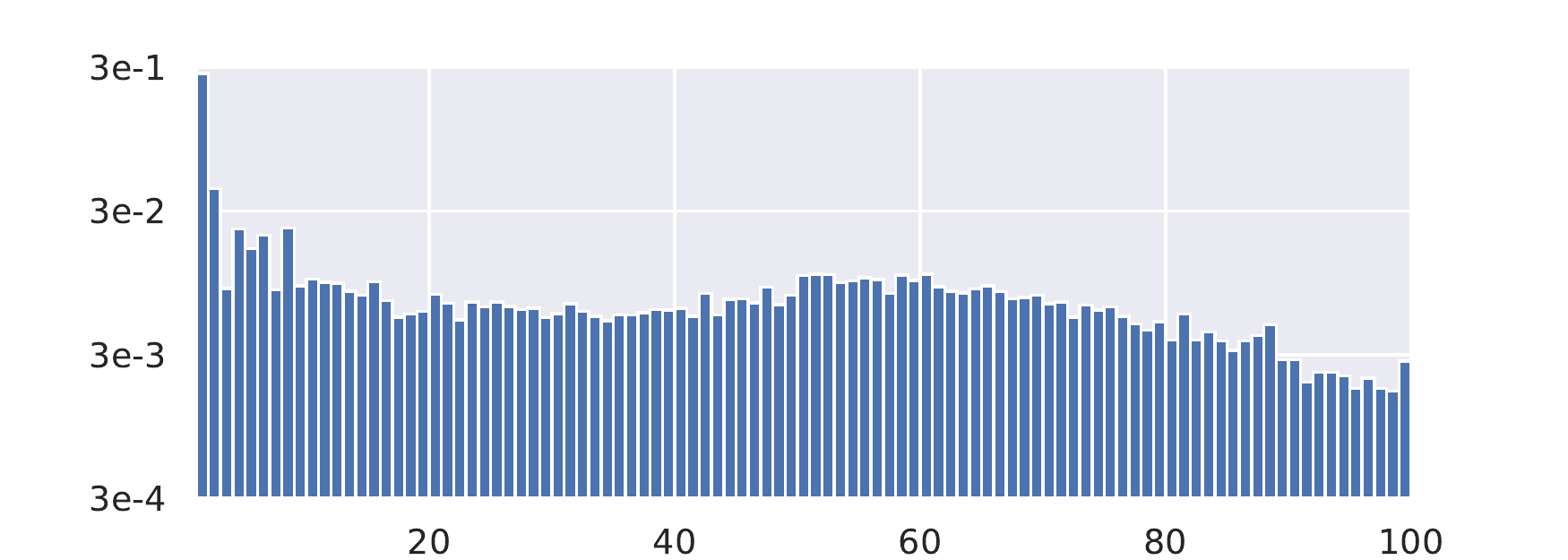}
      \caption{ \bfpmedium ($\mu=29.5$, median: 19)}
  \end{subfigure}
  \caption{Length histograms of \copySpan{\cdot}{\cdot} actions during
      beam decoding in log-$y$ scale. Beam merging is disabled
      for computing the statistics of this experiment. For \bfpsmall 11.2\%
      of the copy actions are single-copy actions, whereas for \bfpmedium
      27.1\% of the actions are single-copy actions. This suggests that
      \seqCopySpan uses long span-copying actions in the majority of
      the cases where it decides to take a span-copying action.}\label{fig:length hists}
\end{figure*}
For a quantitative analysis, we additionally compute statistics for the greedy
decoding strategy of \seqCopySpan.
In \rF{fig:length hists}, we plot the frequency of the lengths of the copied spans for \bfpsmall and \bfpmedium. Given
that the merging mechanism in beam decoding does \emph{not} offer a unique way for measuring
the length of the copied spans (actions of different lengths are often merged), we disable
beam merging for these experiments and employ greedy decoding.
Overall, the results suggest that the model learns to copy long sequences, although
single-copy actions (\eg, to re-use a variable name) are also common.
In the \bfpsmall dataset, \seqCopySpan picks a \copySpan{\cdot}{\cdot} action
with a span longer than one token about three times per example, copying spans
9.6 tokens long on average (median 7).
Similarly in \bfpmedium, \seqCopySpan copies spans of 29.5 tokens long (median 19)
This suggests that the model has learned to take advantage of the span-copying
mechanism, substantially reducing the number of actions that the decoder needs
to perform.

We also find that the \seqSimpleCopy model tends to (mistakenly) assign higher scores to the
input sequence, with the input sequence being predicted as an output more often compared
to the span-copying model: the MRR of the input sentence is 0.74 for the baseline
\seqSimpleCopy model compared to 0.28 for the \seqCopySpan model in the
\bfpsmall dataset.
This suggests that the strong bias towards copying required of the baseline model
(as most of the decoding actions are single-token copies) negatively impacts the
tendency to generate \emph{any} change.
In appendix \ref{app:visualisations:greedy-code-repair}, we include a visualisation of the
attention of a greedy decoding on a single example.

\paragraph{Grammar Error Correction}
A counterpart to code repair in natural language processing is grammar error
correction (GEC).
Again, our span-copying model is a natural fit for this task.
However, this is a rich area of research with highly optimised systems,
employing a series of pretraining techniques, corpus filtering,
deterministic spell-checkers, \etc
As we believe our contribution to be orthogonal to the addition of such
systems, we evaluate it in a simplified setting.
We only compare our \seqCopySpan model to our baseline \seqSimpleCopy
model, expecting results substantially below the state of the art and only
highlighting the \emph{relative} improvement our contribution offers.
Our models have a 2-layer bi-GRU encoder with a hidden size of 64,
a single layer GRU decoder with hidden size of 64, tied embedding layer
of size 64 and use a dropout rate of 0.2.

\begin{table}\centering
    \caption{Evaluation on Grammar Error Correction (GEC) Task~\citep{bryant2017automatic}. Note that our models
    use no pretraining, spell checking or other external data, which are
    commonly used in GEC tasks.} \label{tbl:gec eval}
    \begin{tabular}{@{}lrrr@{}} \toprule
        & Prec. & Recall & F$_{0.5}$ \\ \midrule
\seqSimpleCopy & \textbf{34.9}\% & 6.4\% & 0.1853\\
\seqCopySpan  & 28.9\% & \textbf{10.4}\%  & \textbf{0.2134}\\
    \bottomrule \end{tabular}
\end{table}
We use training/validation folds of the FCE~\citep{yannakoudakis2011new} and W\&I+LOCNESS~\citep{granger1998computer,bryant2019bea}
datasets for training and test on the test fold of the FCE dataset. These datasets
contain sentences of non-native English students along with ground-truth grammar
error corrections from native speakers.
\rTab{tbl:gec eval} shows the results computed with the ERRANT evaluation metric~\citep{bryant2017automatic},
where we can observe that our span-copying decoder again outperforms the
baseline decoder.
Note that the results of both models are substantially below those of state
of the art systems (\eg \citet{grundkiewicz2019neural}), which employ
 (a) deterministic spell checkers
 (b) extensive monolingual corpora for pre-training
 and
 (c) ensembling.

%% file: figures/repairexample.tex
\tikzset{%
highlight/.style={rectangle,fill=green!30,draw,fill opacity=0.25,inner sep=0pt}
}
\newcommand{\tmk}[1]{\tikz[overlay,remember picture,baseline=(#1.base)] \node (#1) {\color{white}I};}
\hspace{0.05\textwidth}
\begin{minipage}{0.45\textwidth}
\textbf{Input}
\begin{lstlisting}[language=Java,basicstyle=\scriptsize\ttfamily]
(*\tmk{cp1Start}*)public boolean equals(Object (*\tmk{cp3Start}*)obj(*\tmk{cp3End}*) ){(*\tmk{cp1End}*)


(*\tmk{cp2Start}~~*) return this.equals(obj);
}(*~~~~~~~~~~~~~~~~~~~~~~~~~~\tmk{cp2End}*)
\end{lstlisting}
\end{minipage}
\hspace{0.10\textwidth}
\begin{minipage}{0.45\textwidth}
\textbf{Output}
\begin{lstlisting}[language=Java,basicstyle=\scriptsize\ttfamily]
(*\tmk{cp1StartTo}*)public boolean equals(Object obj){(*\tmk{cp1EndTo}*)
    if ( (*\tmk{cp3StartTo}*)obj(*\tmk{cp3EndTo}*) == null)
        return false;
(*\tmk{cp2StartTo}~~*)  return this.equals(obj);
}(*~~~~~~~~~~~~~~~~~~~~~~~~~~\tmk{cp2EndTo}*)
\end{lstlisting}
\end{minipage}

\begin{tikzpicture}[overlay,remember picture]
\node[highlight,fit=(cp1Start.north west) (cp1End.south east)] (cp1inp) {};
\node[highlight,fit=(cp1StartTo.north west) (cp1EndTo.south east)] (cp1out) {};
\draw[<-, dashed, black!60!green] (cp1End) -- (cp1StartTo) node [pos=0.5,above] {\scriptsize $a_1$: \copySpan{1}{9}};

\node[highlight,fill=blue!15,fit=(cp3Start.north west) (cp3End.south east)] (cp3inp) {};
\node[highlight,fill=blue!15,fit=(cp3StartTo.north west) (cp3EndTo.south east)] (cp3out) {};
\draw[<-, dashed, blue] (cp3End.south west) to [out=-23, in=169] (cp3StartTo) node {};
\draw[<-, dashed, white] (cp3End) -- (cp3StartTo) node [pos=0.65,below] {\color{blue}\scriptsize $a_4$: \copySpan{6}{7}};

\node[highlight,fill=brown!15,fit=(cp2Start.north west) (cp2End.south east)] (cp2inp) {};
\node[highlight,fill=brown!15,fit=(cp2StartTo.north west) (cp2EndTo.south east)] (cp2out) {};
\draw[<-, dashed, black!60!brown] ($(cp2inp.east) + (0, 0.125)$) -- (cp2StartTo) node [pos=0.5,below] {\scriptsize $a_8$: \copySpan{9}{18}};

\end{tikzpicture}

%% file: text/conclusion.tex
We have presented a span-copying mechanism for commonly used
encoder-decoder models. In many real-life tasks, machine learning
models are asked to edit a pre-existing input. Such models can
take advantage of our proposed model. By correctly and efficiently
marginalising over all possible span-copying actions we can encourage
the model to learn to take a single span-copying action rather than
multiple smaller per-token actions.

Of course, in order for a model to copy spans, it needs to be able
to represent all possible plans which is $O(n^2)$ to the input size.
Although this is memory-intensive, $O(n^2)$ representations are
common in sequence processing models (\eg in transformers). In
the future, it would be interesting to investigate alternative
span representation mechanisms. Additionally, directly optimising
for the target metrics of each task (rather than negative
log-likelihood) can further improve the results for each task.

%% file: text/appendix.tex
\section{Additional Experimental Results}

\subsection{Detailed Fixer Evaluation Results}
\label{app:results:fixer-details}
\rTab{tbl:fixer eval} shows a breakdown of the performance
of our models on the fixers dataset of \citet{yin2019learning}.

\begin{table}[t]\centering \scriptsize
    \caption{C\# Fixer Accuracy (\%) in the One-Shot Generation Task}\label{tbl:fixer eval}
    \begin{tabular}{lrrrrr} \toprule
         & \multicolumn{2}{c}{\seqSimpleCopy}    && \multicolumn{2}{c}{\seqCopySpan}  \\ \cmidrule{2-3} \cmidrule{5-6}
         &  @ 1               & @ 5     && @ 1               & @ 5        \\ \midrule
CA2007   &  16.8  & 24.4    &&  36.9     &  46.5    \\
IDE0004  &  14.8  & 20.8    &&  23.5     &  33.6    \\
RCS1015  &  24.0  & 25.3    &&  23.9     &  26.8    \\
RCS1021  &   1.8  &  4.4    &&  7.8      & 16.8     \\
RCS1032  &   1.8  &  2.7    &&  2.5      &  3.7     \\
RCS1058  &  20.6  & 20.9    &&  19.9     & 22.7     \\
RCS1077  &  3.2   &  3.9    &&  4.5      &  5.8     \\
RCS1089  &  59.8  & 59.8    &&  59.8     & 59.9     \\
RCS1097  &  1.6   & 3.7     &&  14.9     & 27.7     \\
RCS1118  &  45.1 & 69.6     &&  46.0     & 55.6     \\
RCS1123  &  15.8 & 19.5     &&  27.7     & 22.7     \\
RCS1146  &  12.2 & 16.5     &&  19.7     & 31.5     \\
RCS1197  &   1.1 & 1.8      &&   1.7     & 2.3      \\
RCS1202  &  6.5 & 8.4       &&  11.6     & 23.3     \\
RCS1206  &  34.9 & 35.0     &&  36.2     & 37.5     \\
RCS1207  & 2.1  & 4.2       &&  5.0      &  8.2    \\
    \bottomrule \end{tabular}
\end{table}

\subsection{Evaluation of ``Merged'' Beam Search}
\label{app:results:merged-beam-search}

\begin{table*}[t]\centering
    \caption{Evaluation of our beam search method vs. simpler decoding variants on
     the code repair task.}%
    \label{tbl:repairEval-beam-search}
    \begin{tabular}{@{}llrrrr@{}}
      \toprule
        && Accuracy & Accuracy@20 & MRR & Structural Match \\
      \midrule
        \multicolumn{2}{l}{\underline{On \bfpsmall}}\\
        \qquad& \seqCopySpan                     & \textbf{17.7\%} & \textbf{45.0\%} & \textbf{0.247} & \textbf{21.2\%}\\
        \cmidrule(lr){2-6}
        \qquad& \seqCopySpan (merge rays at end)  &         17.5\%  &         41.6\%  &         0.242  & \textbf{21.2\%}\\
        \qquad& \seqCopySpan (greedy decoding)    &         15.3\%  &            ---  &           ---  &         17.9\% \\
        \multicolumn{2}{l}{\underline{On \bfpmedium}}\\
        \qquad& \seqCopySpan                &  \textbf{8.0\%} & \textbf{25.4\%} & \textbf{0.105} & \textbf{13.7\%}\\
        \cmidrule(lr){2-6}
        \qquad& \seqCopySpan (merge rays at end)  &     6.8\%  &         18.0\%  &         0.083  &         11.5\% \\
        \qquad& \seqCopySpan (greedy decoding)    &     1.0\%  &         ---     &            --- &         3.2\%   \\
      \bottomrule
    \end{tabular}
\end{table*}

We show experimental results for our choices in the beam decoding algorithm \rA{alg:beam-search}
in \rTab{tbl:repairEval-beam-search}.
When using a ``standard'' beam decoding algorithm in which we merge the
probabilities of different rays only in a single post-processing step (\ie directly
implementing Eq. \rEq{eq:seq-prob3}), the accuracy of the top beam search result
is only marginally worse, but the accuracy when considering the full beam is
considerably affected.
This is expected, as \rA{alg:beam-search} merges rays earlier,
``freeing up'' space in the beam for other results and indicates that the added
computational effort for merging beams allows the model
to generate more diverse hypotheses.

Similarly, using a greedy decoding mechanism alone is doing substantially worse,
as expected, especially on the \bfpmedium dataset, where the greedy decoder
often fails to produce meaningful output.

\section{Visualisations}

\subsection{Visualisation of Span-Copying Attention}
\label{app:visualisations:greedy-code-repair}
\rF{fig:copyspanviz} visualises the copy-span attention for
the greedy action sequence for the example in \rF{fig:code repair example}.

\begin{figure*}
    \input{figures/copyspanattention.tex}
    \caption{Visualisation of the attention weights
        over some of the greedy action sequence of \rF{fig:code repair example}.
        To fit this figure within one page, we only visualise
        attention at $a_1$, $a_2$, $a_3$, $a_4$, $a_5$, $a_8$.
        Note that the colour range changes per-figure to allow for a better
        contrast in the visualisation. Best viewed in screen.}\label{fig:copyspanviz}
\end{figure*}

%% file: figures/copyspanattention.tex
\centering
\begin{subfigure}[b]{0.46\textwidth}\centering
        \includegraphics[width=0.85\textwidth]{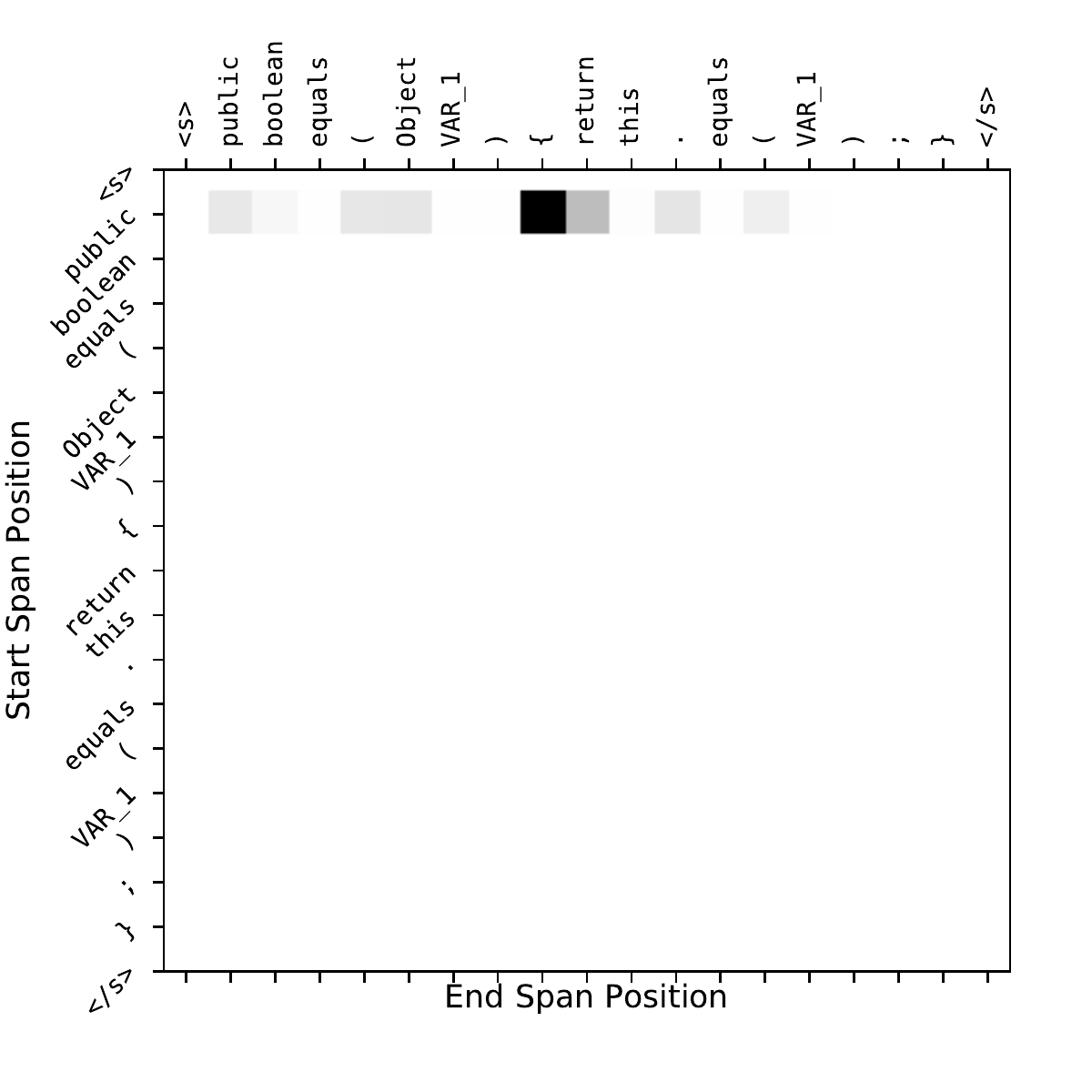}
        \caption{$a_1$: \copySpan{1}{9} has a probability of 42.8\%. The
            model is also predicting \genTok{public} with 0.4\% probability.}
\end{subfigure}
~
\begin{subfigure}[b]{0.46\textwidth}\centering
    \includegraphics[width=0.85\textwidth]{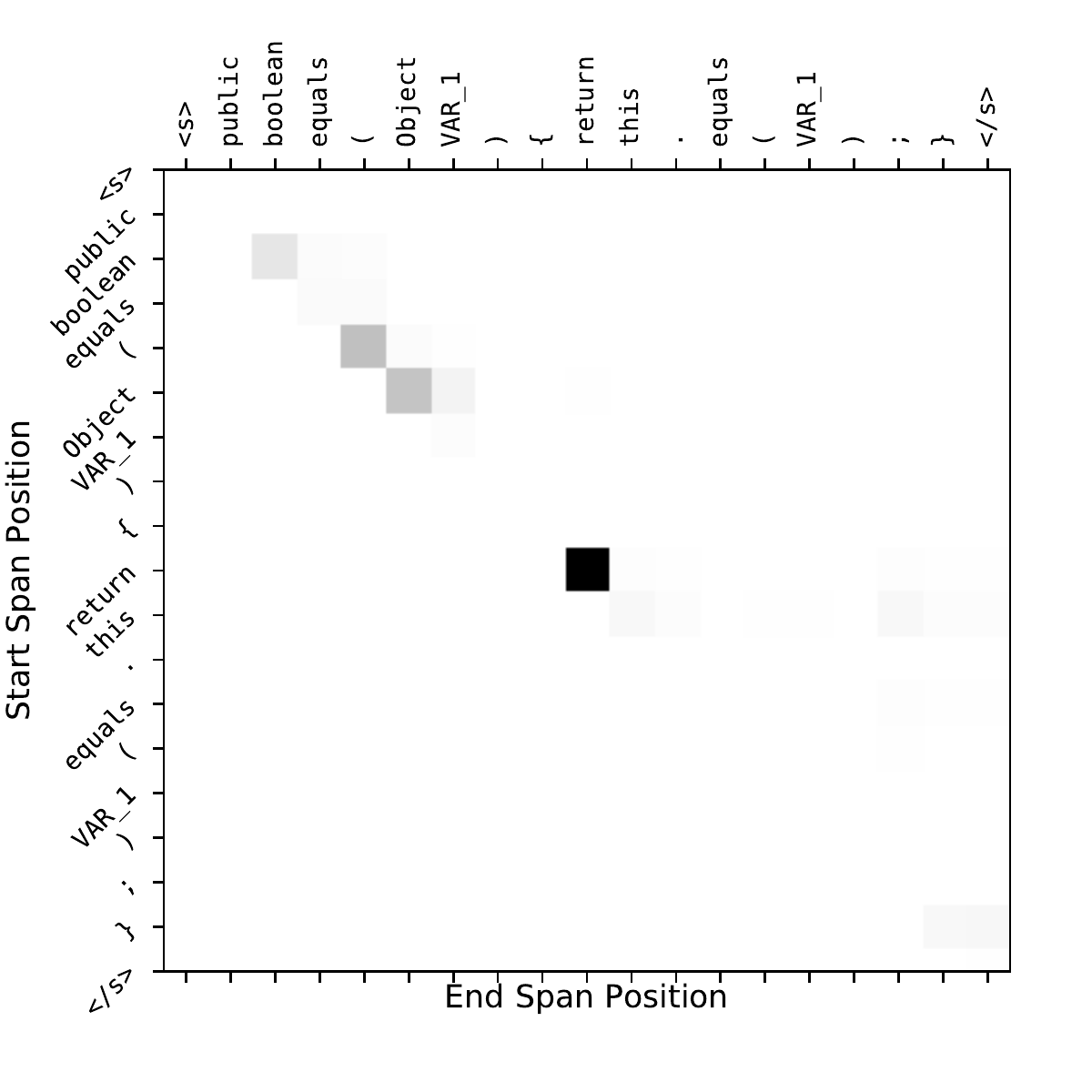}
    \caption{$a_2$: \genTok{\code{if}} has probability 59.3\% where as the
        highest span-copying action (\copySpan{9}{10}) has a probability
        of only 0.8\%.}
\end{subfigure}

\begin{subfigure}[b]{0.46\textwidth}\centering
    \includegraphics[width=0.85\textwidth]{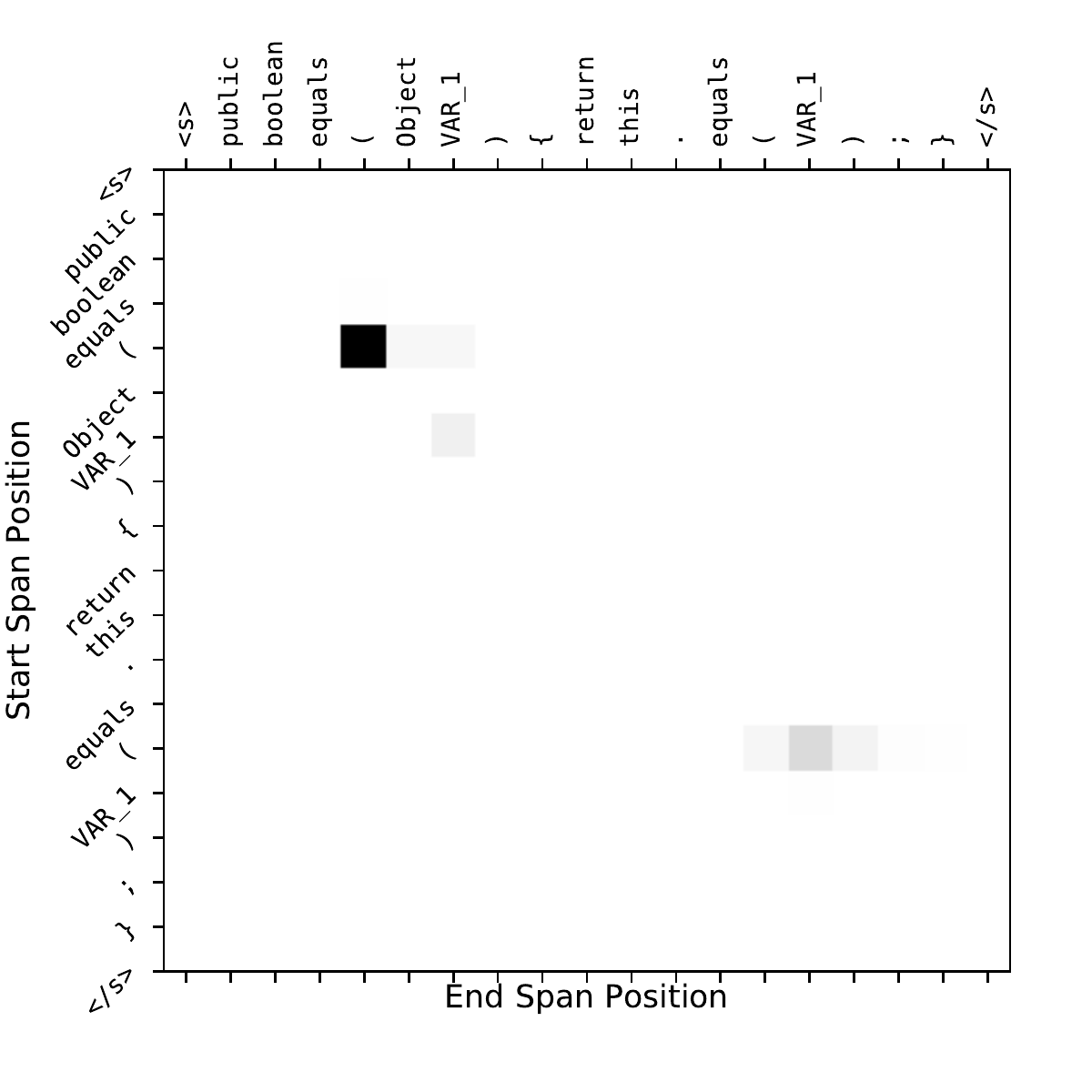}
    \caption{$a_3$: \genTok{\code{(}} has a probability of 99.9\%. The highest
        span-copying action is the correct \copySpan{4}{5}
        but with negligible probability.}
\end{subfigure}
~
\begin{subfigure}[b]{0.46\textwidth}\centering
    \includegraphics[width=0.85\textwidth]{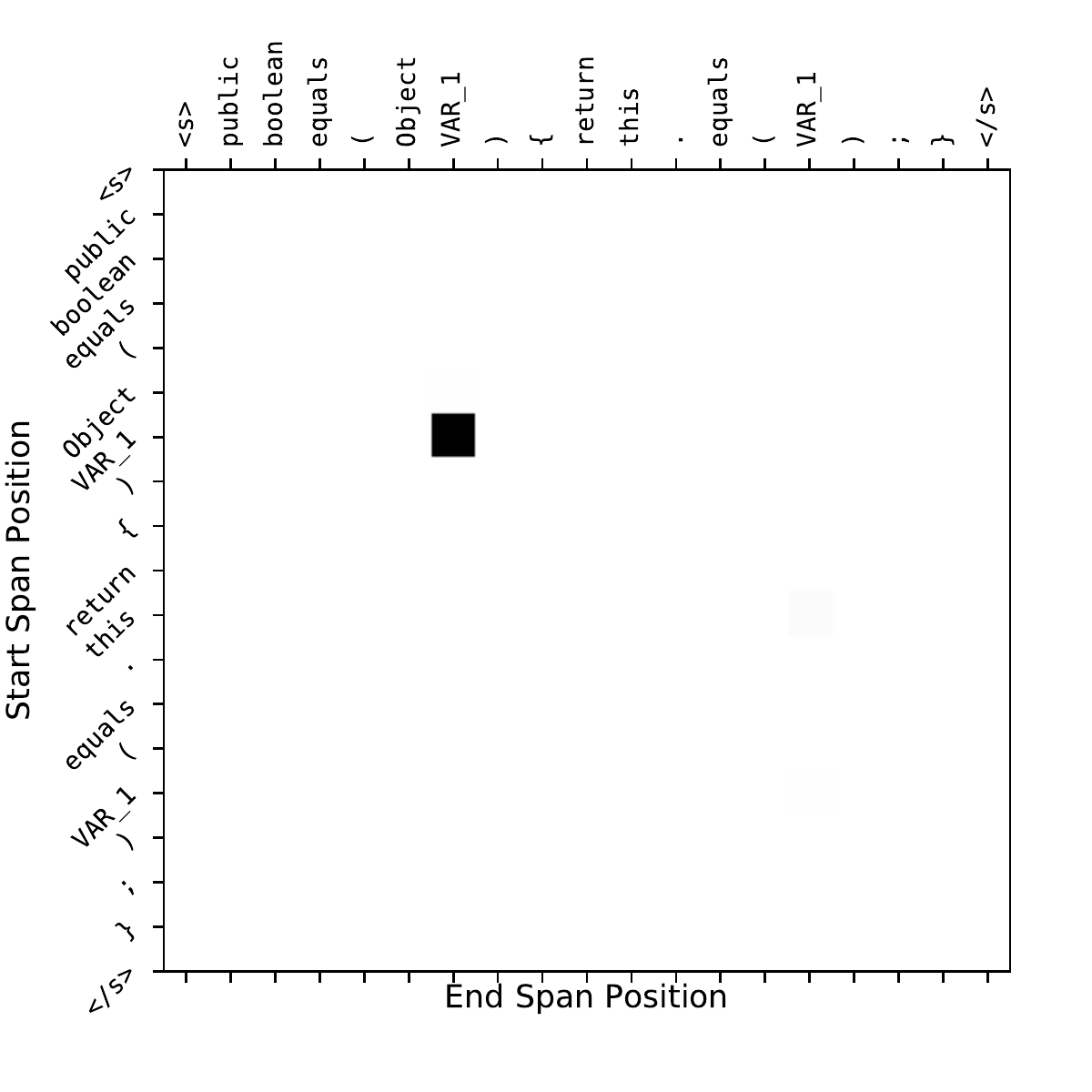}
    \caption{$a_4$: \copySpan{6}{7} has a probability of 69.9\%. The
    model is also (mistakenly) assigning a 24.3\% probability to  \genTok{\code{(}}.}
\end{subfigure}

\begin{subfigure}[b]{0.46\textwidth}\centering
    \includegraphics[width=0.85\textwidth]{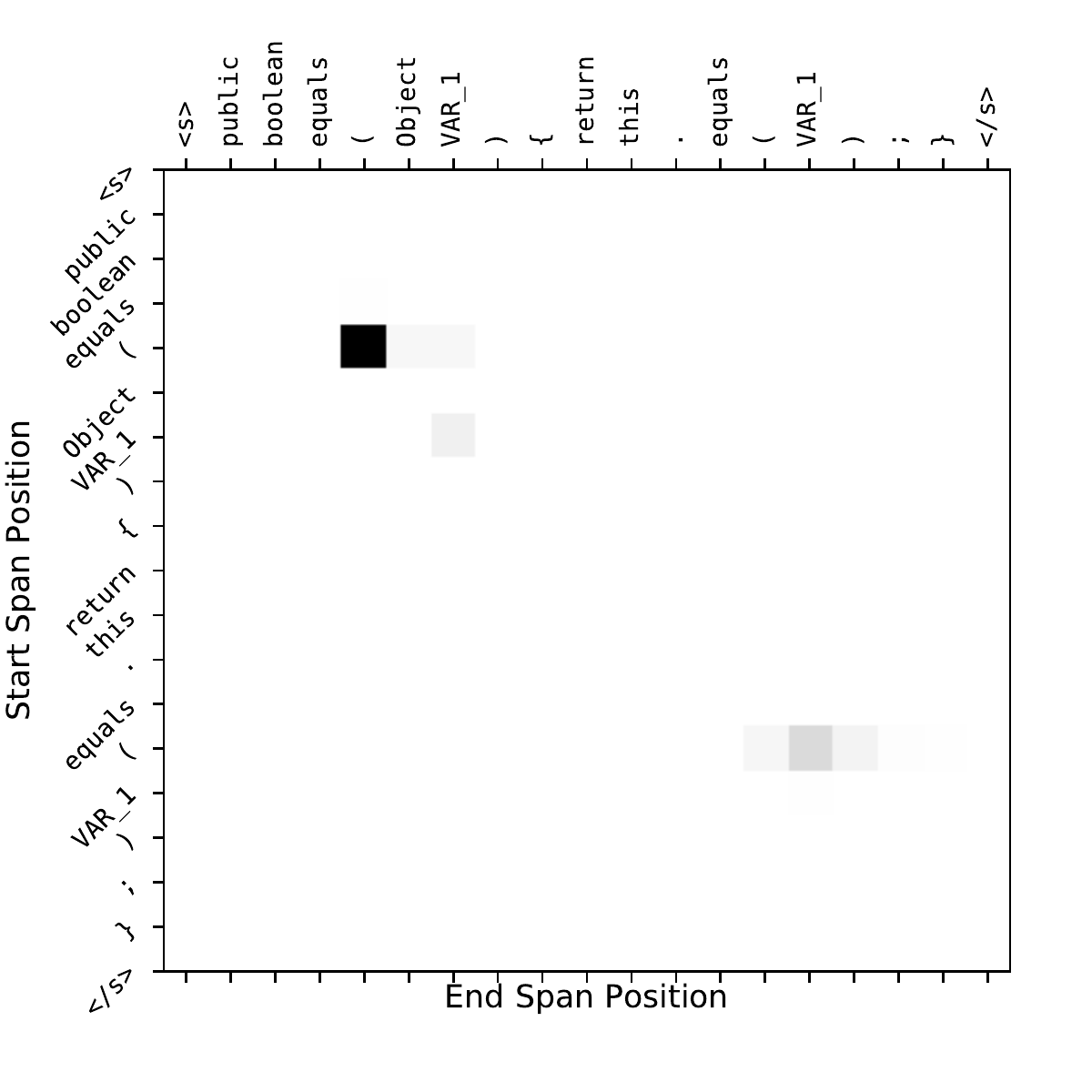}
    \caption{$a_5$: \genTok{\code{==}} has a probability of 66.0\%. The highest
        span-copying action is \copySpan{7}{8} but with 0.8\% probability.}
\end{subfigure}
~
\begin{subfigure}[b]{0.46\textwidth}\centering
    \includegraphics[width=0.85\textwidth]{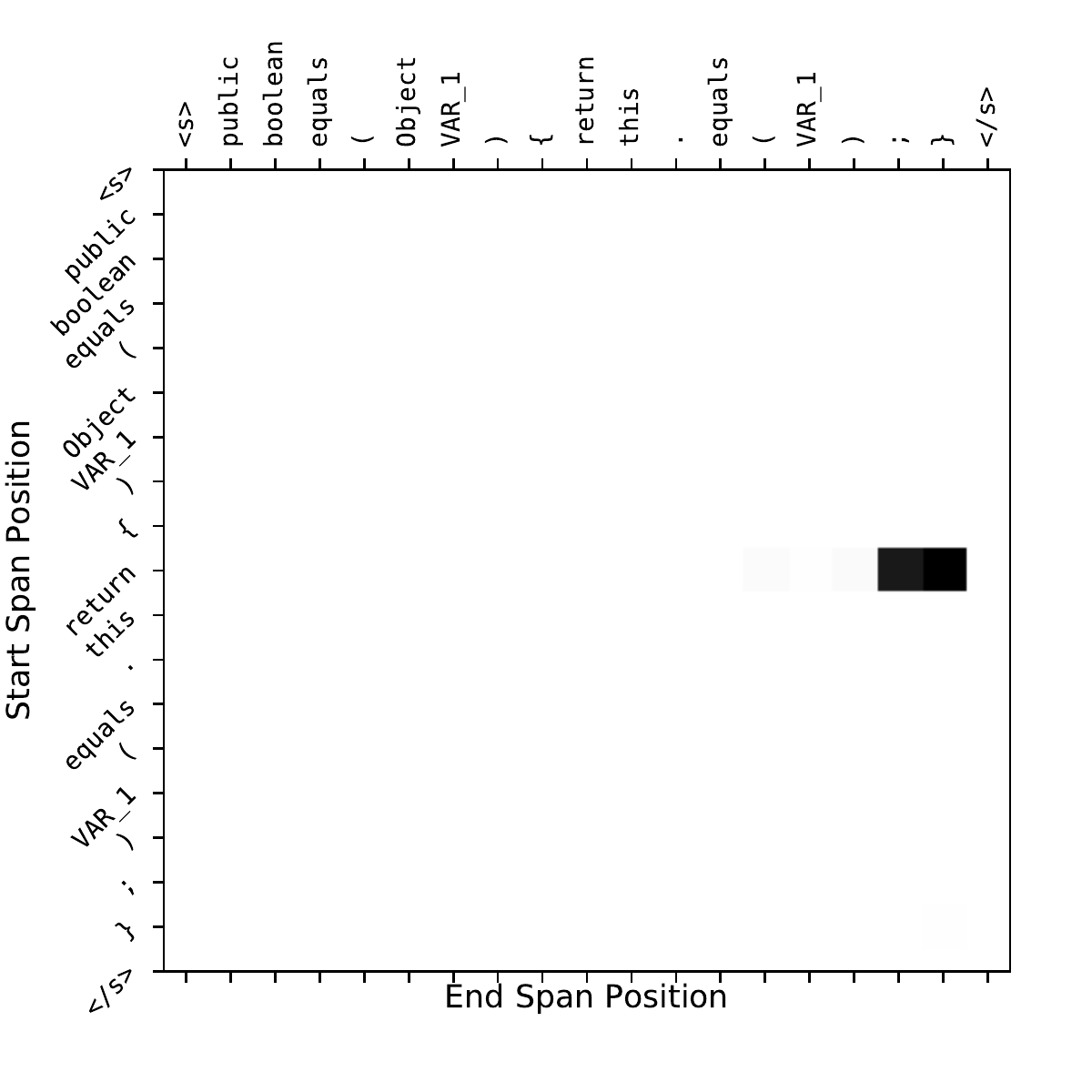}
    \caption{$a_8$: \copySpan{9}{18} has a probability of 43.7\%. The
    model is also (mistakenly) assigning a 8.3\% probability to  \genTok{\code{else}}.}
\end{subfigure}